\documentclass[11pt]{article}
\usepackage{nodalida2025}
\usepackage{times}
\usepackage{url}
\usepackage{latexsym}
\usepackage{graphicx}
\usepackage{hyperref} 
\usepackage[T1]{fontenc}

\aclfinalcopy % Uncomment this line for the final submission

\title{Aligning Language Models for Icelandic Legal Text Summarization}

\author{Þórir Hrafn Harðarson \\
  Department of \\Computer Science \\
  Reykjavik University \\
  Iceland\\
  {\tt thorirhh21@ru.is} \\\And
  Hrafn Loftsson \\
  Department of \\Computer Science \\
  Reykjavik University \\
  Iceland\\
  {\tt hrafn@ru.is} \\\And
  Stefán Ólafsson \\
  Department of \\Computer Science \\
  Reykjavik University \\
  Iceland\\
  {\tt stefanola@ru.is } \\}

\date{}

\begin{document}
\maketitle
\begin{abstract}
    The integration of language models in the legal domain holds considerable promise for streamlining processes and improving efficiency in managing extensive workloads. However, the specialized terminology, nuanced language, and formal style of legal texts can present substantial challenges. This study examines whether preference-based training techniques, specifically Reinforcement Learning from Human Feedback and Direct Preference Optimization, can enhance models' performance in generating Icelandic legal summaries that align with domain-specific language standards and user preferences. We compare models fine-tuned with preference training to those using conventional supervised learning. Results indicate that preference training improves the legal accuracy of generated summaries over standard fine-tuning but does not significantly enhance the overall quality of Icelandic language usage. Discrepancies between automated metrics and human evaluations further underscore the importance of qualitative assessment in developing language models for the legal domain.
\end{abstract}

% KEYWORDS

% Icelandic, Large Language Models, Text Summarization, Legal Domain, Domain-Specific Language, Reinforcement Learning from Human Feedback , Direct Preference Optimization

\section{Introduction}

The development of language models (LMs) for use in specialized, professional domains has the potential to create time-saving, value-adding processes. This may benefit various fields such as law, healthcare, and engineering, where much of the work involves analyzing and writing domain-specific texts and documents. 

This is particularly relevant in the legal domain. An analysis of the legal systems in the USA and Germany from 1998 to 2019 reported a monolithic growth in these systems \cite{Coupette2021}. Massive volumes of text data are a byproduct of most modern legal systems \cite{Katz_2020}, leading to an environment with an ever-increasing amount of source material. Consequently, lawyers and attorneys must devote more time to analyzing and reviewing legal documents while preparing their casework, resulting in a growing workload in an already overburdened profession \cite{logmannabladid, nickum2023burnout}.

A comprehensive awareness and understanding of relevant laws and precedents is paramount to success in legal arguments. Therefore, the ability to quickly summarize legal sources may significantly reduce the time spent reviewing pertinent material \cite{jain2021legal_summarization}. Summaries can also serve as references for justifying claims and building cases. This is an area where generative LMs can be particularly useful, by processing and analyzing the bulk of the text needed. 

In Iceland, there are substantial requirements within the legal domain that the quality of text meets the linguistic standards of the domain, both in terms of domain-specific terminology and general Icelandic language proficiency. Consequently, LMs must adhere to the professional standards of the domain in which they are applied. The legal domain is also characterized by a specialized vocabulary, particularly formal syntax, and semantics based on extensive domain-specific knowledge \cite{legal_language}. This makes the task of aligning LMs to the specific language of the legal domain a non-trivial issue.

The most common method to enhance the capabilities of a pre-trained generative LM is instruction fine-tuning, where the model receives an instruction as input and the correct response  as the target label \cite{gpt_paper,liu2019roberta}. Under this paradigm, the model is rewarded for correctly following the instructions; however, this does not necessarily entail that it captures the linguistic nuances within the target texts. \textit{Reinforcement Learning from Human Feedback} (RLHF) is one such method that uses algorithms and reward-based methods from reinforcement learning (RL) to directly optimize a LM based on data collected from human feedback \cite{stiennon2022learning}, aiming to help the model align better with both subjective and complex texts. Another more recent approach, based on the same principle, is \textit{Direct Preference Optimization} (DPO) \cite{dpo_paper}, which optimizes the model by transforming the RL reward maximization problem into a more simple classification problem. Though the complexity of the DPO method is less than that of RLHF, it is unclear which method is best suited to align LMs for summarizing Icelandic legal text.

This paper addresses the following research question:

\paragraph{RQ:} Can preference training methods, such as DPO and RLHF, enhance the ability of LMs to generate domain-specific Icelandic texts that users prefer, compared to LMs fine-tuned solely with supervised learning?\\

We compared the quality of text summaries generated for the Icelandic legal domain by models fine-tuned with preference training to those fine-tuned solely through supervised learning. Our findings indicate that applying either RLHF or DPO on top of domain-specific pre-training and instruction fine-tuning can improve the legal accuracy of the generated summaries. However, no similar improvements were observed in the general quality of Icelandic language usage. Additionally, there were discrepancies between automated numerical evaluations and qualitative human assessments.

\section{Background and Related Work}

Transformer-based language models (LMs) have become central to text generation and NLP tasks, largely due to their adaptability when fine-tuned on specific tasks \cite{vaswani2017attention, wolf2020transformers}. These models, typically containing billions of parameters \cite{llama2}, excel at few-shot or zero-shot tasks that previously required supervised fine-tuning \cite{brown2020language}. However, languages with smaller speaker populations, such as Icelandic, face challenges due to limited representation in training data. Efforts to address this include IceBERT, a masked LM for Icelandic \cite{snaebjarnarson-etal-2022-warm}, and GPT-SW3, a multilingual model covering most Nordic languages \cite{gptsw3}. These initiatives align with ongoing government initiatives in Iceland to preserve the Icelandic language midst the rapid advancements in language technology \cite{nikulasdottir-etal-2020-language} and with the government's partnership with OpenAI. \footnote{\url{https://openai.com/index/government-of-iceland/}}

Recent LM advancements emphasize RLHF to improve performance. Initial work by OpenAI explored human feedback to refine RL reward functions for complex tasks \cite{christiano2023deep}. \citet{stiennon2022learning} applied RLHF in NLP, training models for improved text summaries. \citet{ouyang2022training} extended this approach with InstructGPT, producing outputs that were preferred over those from larger models like GPT-3. RLHF-trained models have shown advantages in common sense reasoning and world knowledge \cite{glaese2022improving}. A more streamlined approach, Direct Preference Optimization (DPO), optimizes the model directly via preference-based comparisons, showing similar performance to RLHF with faster results \cite{dpo_paper, tunstall2023zephyr}.

Given the powerful text processing capabilities of modern LMs, numerous studies have explored their applications in the legal domain, including judgment prediction \cite{trautmann2022legalpromptengineeringmultilingual}, statutory interpretation \cite{blairstanek2023gpt3performstatutoryreasoning}, legal reasoning \cite{yu2022legalpromptingteachinglanguage}, and using large models like ChatGPT as proxy legal advisors \cite{pettinato2023chatgpt}. Research has also assessed performance on legal exams to gauge legal reasoning capabilities \cite{choi2023chatgpt}.

For domain-specific improvements, LEGAL-BERT \cite{chalkidis2020legalbertmuppetsstraightlaw} demonstrates the advantages of pre-training a LM specifically for legal tasks, finding that additional domain-specific pre-training on legal corpora improved performance compared to using general-purpose BERT. Building on this work, \citet{LICARI2024105908} developed Italian LEGAL-BERT, which they used in experiments for legal text summarization \cite{italian_legal_bert_summaries}. Another Italian research, The PRODIGIT Project \cite{pont2023legalsummarisationllmsprodigit}, is a large-scale initiative aiming to support tax lawyers by utilizing LMs for summarization. In a similar line of work, \citet{schraagen2022abstractive} applied a BART-based LM for summarization of Dutch case verdicts. The LM-generated summaries were considered useful in human evaluations, although they still fall short of the quality of human-generated summaries.

\section{Methods}
\label{sec:methods}
We selected two open-source models for experimentation in generated Icelandic legal summaries. The first model, a 1.3B parameter version of GPT-SW3, has been pre-trained on Nordic languages using the Nordic Pile, a large corpus of approximately 1.2 TB, containing data in Swedish, English, Norwegian, Danish, and Icelandic \cite{gptsw3, öhman2023nordic}. The second model was a 7B parameter version of Llama2 \cite{llama2}, mostly pre-trained on English texts.\footnote{Llama2 was the most powerful available open source models at the time of experimentation for this research.} With this setup, we compared the effectiveness of language-specific pre-training (GPT-SW3) to the general learning capacity of a larger model (Llama2).

To better understand the effect of pre-training on Icelandic texts, we created a sub-corpus of the Icelandic Gigaword Corpus (IGC) \cite{igc_four_versions,igc_risamalheild} that contained 10\% of its data sampled at random. We then created a version of Llama2 (called Ice-Llama2) that was pre-trained on this sub-corpus. 

All models were trained in three phases. In the first phase, the models were further pre-trained on domain-specific Icelandic legal text (see Section \ref{sec:datasets}) and in the second phase, the models were fine-tuned to perform the supervised court case summarization task. After this training phase, the model able to produce the highest ROUGE score \cite{lin2004rouge} -- a commonly used metric for summarization tasks -- was used to create summaries for a pairwise comparison dataset. Finally, in the third phase, this data was then used to perform preference training with DPO and RLHF.

\subsection{Datasets}
\label{sec:datasets}
The datasets used for the training process are based on case rulings from the Icelandic supreme court, publicly available on the court's website\footnote{\url{https://www.haestirettur.is/domar/}}. One row of data consists of a court ruling and a summary made by a lawyer or attorney.

Many of the court rulings are too long to fit the context window of the chosen models. We therefore split the data into two parts: 1) long court rulings only (R); 2) court rulings that fit the window and their summaries (RS). The R dataset was used for the first phase of training, namely for further pre-training the models on domain-specific Icelandic legal text. The RS dataset was thus used for the second phase, fine-tuning the models to perform the summarization task.
After splitting the data in this manner, the R dataset contained 5677 rows, split into 5077 rows (90\%) of training data, 300 rows (5\%) of test data and 300 rows (5\%) of validation data. This left the RS dataset with 2,613 rows of data, further split into 2013 rows (78\%) of training data,  300 rows (11\%) of test data and 300 rows (11\%) of validation data\footnote{\url{https://huggingface.co/datasets/thorirhrafn/domar_data}}.

\subsection{Domain Specific Further Pre-Training}

To investigate the importance of further pre-training on domain-specific  text, the models were trained on the R dataset of court rulings only. As auto-regressive models, they were trained using self-supervised learning by shifting the input sequence forward by one token, creating target labels for predicting the next token in the sequence. The legal text in the dataset was processed by packing chunks of text together and dividing them into fixed-size blocks of 512 tokens. To measure the improvement in domain-specific text generation, the perplexity of both models on Icelandic legal text was estimated before and after fine-tuning, and the results were compared.

\subsection{Instruction Fine-Tuning}

Following the domain-specific training step, the models were fine-tuned using supervised learning to generate summaries. This was done using the RS dataset, where the models were given an input consisting of an instruction to create a summary, followed by the ruling text, and a token to mark the start of the summary. The corresponding label was the human-generated summary of the ruling from the court's website. Due to the sequence length, the data was fed to the models in mini-batches of single rulings, but to ensure more stable training, the models processed eight sequences before calculating the gradient and updating the weights. To evaluate the models' ability to generate summaries and the impact of supervised fine-tuning, the ROUGE score was computed both before and after training.

\subsection{Preference Training}

The third phase of training was to apply preference training on top of the instruction fine-tuning to determine if it would improve performance. DPO requires a specialized dataset where the model is presented with two responses: one marked as preferred and the other as rejected. Then, it uses a loss function to compare these responses, directly penalizing the model for generating outputs that resemble the rejected data, increasing the likelihood of the model producing outputs that align with the preferred responses.

Implementing RLHF first involves training a reward model that serves as a reward function during training by classifying generated summaries and assigning scalar values based on its evaluation. This reward model was fine-tuned using a binary classification task on the dataset of preferred and rejected responses, which was also used for the DPO training. Training was carried out using Proximal Policy Optimization (PPO), a policy gradient algorithm that directly optimizes the policy guiding the model’s behavior. The goal is to maximize the probability of actions (i.e., generating summaries) that yield high rewards from the environment, given the current state. PPO limits the policy changes allowed at each training step, thereby ensuring greater stability and improving convergence to an optimal solution. Care must be taken that the values produced by the reward model need to be scaled appropriately. If the reward model's interpretation of preferences is inconsistent or inaccurate, it can produce unstable reward signals, leading to conflicting feedback which can cause divergence during training \cite{stiennon2022learning}.

As before, performance was assessed by calculating the ROUGE score both before and after the RLHF training.

\subsection{Human Evaluation}

The final evaluation involved having legal experts rank the generated summaries. Using the test split of the RS dataset, the trained models were used to generate summaries which were then presented to human experts for ranking. One primary legal expert, an attorney with over five years experience and that has proceeded dozens of court cases, ranked summaries generated from 25 court rulings, selected to represent a wide variety of cases. To assess agreement, two additional legal experts ranked summaries for five of these cases. The primary expert also evaluated each generated summary by assigning two separate scores, each on a scale of 1 to 5: one score for the quality of the summary as a legal text, and another for the quality of the Icelandic used in the generated text. Here, the scores represent the quality expected within the legal domain, with a score of 5 meaning complete legal accuracy, and a near perfect use of the Icelandic language. A score of 1 would indicate a total misunderstanding of the legal argument and a totally unacceptable quality of Icelandic.

\section{Results}

To optimize training efficiency and make the best use of available resources, all models were trained using Low-Rank Adapters (LoRA) \cite{hu2021lora}. The first parameter to be tuned and analyzed during pre-training was the adapter rank value. During the first phase of further pre-training on the R dataset (Icelandic court rulings only), increasing the rank consistently led to a lower loss. This suggests that increasing the number of trainable parameters helps the model to learn better from the training data.

Based on these findings, a relatively large adapter with a rank of 1024 was used for training Llama2-7B on the IGC sub-corpus data and a rank of 256 for phase one (further pre-training on the R data, court rulings only), and a rank of 128 for phase two (training for the summarization task on the RS data, court rulings and summaries).

To assess the impact of phase one, all models were evaluated by calculating their perplexity scores on the test split of the dataset.

\begin{table}[h]
\centering
    \begin{tabular}{|c|c|}
    \hline
    \bf Model & \bf Perplexity \\ \hline
    GPT-SW3-1.3B & 5.281 \\ \hline
    Llama2-7B & 9.283 \\ \hline
    Ice-Llama2-7B &  5.048  \\ \hline
    \end{tabular}
\caption{ Perplexity evaluation on legal text in Icelandic before using further pre-training on legal data. }
\label{tab:tbl_ppl_legal1}
\end{table}

As shown in Table \ref{tab:tbl_ppl_legal1}, the base Llama2 model initially scored significantly higher in perplexity compared to both GPT-SW3 and Ice-Llama2, which had been pre-trained on the ICG sub-corpus data. After the phase one training process, however, both Llama2 variants achieved lower perplexity scores than GPT-SW3.

\begin{table}[h]
\centering
    \begin{tabular}{|c|c|}
    \hline
    \bf Model & \bf Perplexity \\ \hline
    GPT-SW3-1.3B & 4.844 \\ \hline
    Llama2-7B & 2.981 \\ \hline
    Ice-Llama2-7B & 2.900  \\ \hline
    \end{tabular}
\caption{ Perplexity evaluation on legal text in Icelandic after further pre-training on Icelandic court rulings data.}
\label{tab:tbl_ppl_legal2}
\end{table}

\subsection{Instruction Fine-Tuning}

The second phase of training was supervised instruction fine-tuning using the RS dataset (see Section \ref{sec:datasets}), along with the corresponding instruction text and summaries. To determine the optimal number of training epochs, the GPT-SW3 1.3B model was trained for 1, 3, and 5 epochs, with performance evaluated using 10 summaries from the validation set. The model trained for 5 epochs achieved the highest ROUGE scores, so all models were subsequently trained for 5 epochs on the training split of the dataset.

After fine-tuning, the models were evaluated by generating summaries for all 300 entries in the test set. The generated summaries were compared to human-generated baselines using ROUGE scores. As shown in Table \ref{table:rouge_sft_results}, both Llama2-7B variants achieved higher scores than GPT-SW3-1.3B:

\begin{table}[h]
\centering
    \begin{tabular}{|c|c|c|c|}
    \hline
    \bf Model & \bf Rouge1 & \bf Rouge2 & \bf RougeL \\ \hline
    GPT-SW3-1.3B & 0.2829 & 0.1136 & 0.1796 \\ \hline
    Llama2-7B & 0.3055 & 0.1112 & 0.1872 \\ \hline
    Ice-Llama2-7B & 0.3005 & 0.1121 & 0.1861 \\ \hline
    \end{tabular}
\caption{\label{font-table} ROUGE-score evaluation using all 300 summaries in the test dataset after further pre-training on legal data and instruction fine-tuning.}
\label{table:rouge_sft_results}
\end{table}

\subsection{Preference Training}
\subsubsection{Direct Preference Optimization}

In the third phase of training, investigating the impact of additional preference training, we first looked at using the DPO method. Since both Llama2-7B variants achieved nearly identical ROUGE scores after instruction fine-tuning, further training was only applied to the base Llama2-7B model and GPT-SW3-1.3B to highlight the differences between larger models and those with extensive language-specific pre-training.

Following the method used by \citet{tunstall2023zephyr} for training the Zephyr 7B model with DPO, all models were initially fine-tuned for 1, 3, and 5 epochs and then further trained with DPO for an additional 1, 2, and 3 epochs. The resulting ROUGE scores were evaluated to assess the effect of preference training on top of varying levels of supervised fine-tuning. GPT-SW3-1.3B achieved its best improvements after 2 epochs of DPO training, following 5 epochs of fine-tuning. However, performance plateaued after 3 epochs, and in some cases began to degrade, likely due to overfitting.

For the Llama2-7B model, the best improvements were observed after 2 epochs of DPO training but with only 1 epoch of prior fine-tuning. Given the larger parameter count and higher learning capacity of Llama2-7B, the risk of overfitting was more pronounced. To mitigate this, training was conducted with a low starting learning rate of \(7~e-07\), as even slight increases led to overfitting.

After completing this training process, the best-performing versions of GPT-SW3-1.3B and Llama2-7B were evaluated on the entire test set of summaries. This resulted in significant improvements for GPT-SW3-1.3B and modest gains for Llama2-7B, as shown in Table \ref{table:dpo_final_results}:

\begin{table}[h]
\centering
    \begin{tabular}{|c|c|c|c|c|}
    \hline
    \bf Model & \bf Rouge1 & \bf Rouge2 & \bf RougeL \\ \hline
    GPT-SW3-1.3B & 0.3381 & 0.1637 & 0.2263 \\ \hline
    Llama2-7B & 0.3143 & 0.1226 & 0.1963 \\ \hline
    \end{tabular}
\caption{\label{font-table} ROUGE-score evaluation on generating 300 summaries for court rulings in Icelandic after further pre-training on legal data, supervised instruction fine-tuning, and DPO.}
\label{table:dpo_final_results}
\end{table}

\subsubsection{Reinforcement Learning from Human Feedback}

The second preference training method evaluated was RLHF. As outlined in Section \ref{sec:methods}, RLHF involves first training a reward model to classify the output of the policy model and return a scalar reward based on the likelihood that the generated output aligns with human preferences. The same pairwise dataset used during the DPO training phase was utilized to train this reward model. Initial attempts revealed a high susceptibility to overfitting, necessitating training for just a single epoch with a relatively low learning rate of \( 2e^{-06}\).

Early efforts to use the reward model to train a policy using PPO resulted in highly unstable training. The policy quickly learned to exploit the reward model by generating sequences of empty lines, random characters, or incomplete word endings, leading to a spike in KL divergence and inflated rewards. 

 To stabilize the training, the output of the reward model was normalized such that the rewards had a mean of \(\mu=0\) and a standard deviation of \( \sigma=1 \) at the start of training. This normalization led to a much more stable training process. However, the PPO algorithm’s conservative policy updates resulted in slow learning progression, as shown in Figure \ref{fig:rlhf_training}: 

\begin{figure}[htbp]
    \centering
    {\includegraphics[width=0.8\linewidth]{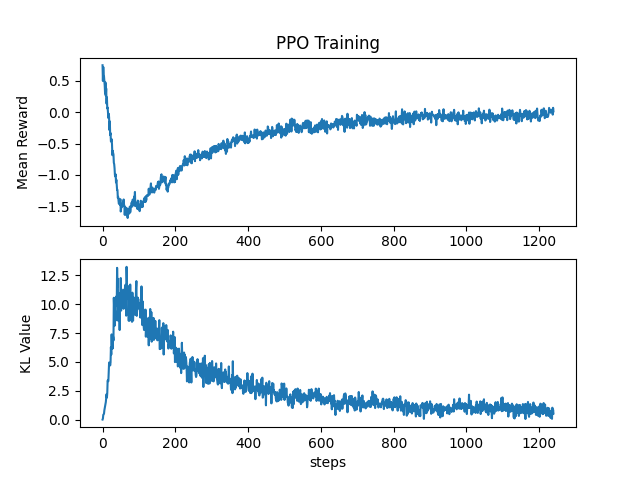}}
    \caption{Mean reward  and KL-divergence for GPT-SW3 1.3B after 20 epochs of training using the PPO reinforcement learning algorithm.}
    \label{fig:rlhf_training}
\end{figure}

Due to the substantial GPU resources required for RLHF, this training method was only applied to the smaller GPT-SW3 model. As with previous evaluations, the model's performance was assessed by calculating the ROUGE score. However, in contrast to its DPO-trained counterpart, the RLHF model did not exhibit performance improvements, as can be seen in Table \ref{table:rlhf_rouge}:

\begin{table}[h]
\centering
    \begin{tabular}{|c|c|c|c|c|}
    \hline
    \bf Model & \bf Rouge1 & \bf Rouge2 & \bf RougeL \\ \hline
    GPT-SW3-1.3B & 0.2690 & 0.1058 & 0.1769 \\ \hline
    \end{tabular}
\caption{\label{font-table} ROUGE-score evaluation on generating 300 summaries for court rulings in Icelandic after further pre-training on legal data, instruction fine-tuning, and reinforcement learning.}
\label{table:rlhf_rouge}
\end{table}

\subsection{Human Evaluation}

As a final evaluation step, the results generated from 25 court rulings by five model variations were ranked by a human expert in the legal domain, tasked with ranking the summaries from $1^{st}$ place to $5^{th}$:

\begin{table}[h]
\centering
    \begin{tabular}{|c|c|c|}
    \hline
    \bf Model & \bf Order & \bf Average \\ \hline
    GPT-SW3-RLHF & 1 & 2.20 \\ \hline
    GPT-SW3-SFT & 2 & 2.36 \\ \hline
    Llama2-DPO & 3 & 3.24 \\ \hline
    Llama2-SFT & 4 & 3.52 \\ \hline
    GPT-SW3-DPO & 5 & 3.68 \\ \hline
    \end{tabular}
\caption{\label{font-table} Average rank for five model variations after being ranked on summary generation for 25 court rulings by a legal expert. The model names have an ending that marks if they were additionally fine-tuned using either DPO or RLHF, or if they were only instruction fine-tuned using supervised learning (SFT).}
\label{table:model_ranking1}
\end{table}

As can be seen in Table \ref{table:model_ranking1}, the version of GPT-SW3-1.3B that had only been instruction fine-tuned and the version that had also been additionally trained with RLHF were most often chosen as the preferred models, despite having achieved the lowest scores during evaluation. Two other experts also ranked the first five of the 25 chosen rulings to get an assessment on the agreement between human legal experts, the results of which can be seen in Table \ref{table:model_ranking2}: 

\begin{table}[h]
\centering
    \begin{tabular}{|c|c|c|c|c|}
    \hline
    \bf Model & \bf Primary & \bf Comparison \\ \hline
    GPT-SW3-SFT & 1.6 & 2.3 \\ \hline
    GPT-SW3-RLHF & 2.2 & 2.6 \\ \hline
    Llama2-DPO & 3.0 & 3.0 \\ \hline
    Llama2-SFT & 4.0 & 4.0 \\ \hline
    GPT-SW3-DPO & 4.2 & 3.2 \\ \hline
    \end{tabular}
\caption{\label{font-table} Average rank for five model variations after being ranked on summary generation for the first 5 court rulings from the list of 25. Rank scores from the primary expert compared to the average from two other legal experts to assess agreement.}
\label{table:model_ranking2}
\end{table}

To further assess the models' capabilities, the primary evaluator assigned each model a score on a scale of 1 to 5, where 1 represented the lowest performance and 5 the highest. The models were evaluated based on two criteria: the quality of the Icelandic language used in the generated text and the legal accuracy in relation to the court ruling being summarized.

\begin{table}[h]
\centering
    \begin{tabular}{|c|c|c|c|c|}
    \hline
    \bf Model & \bf Icelandic & \bf Legal Accuracy \\ \hline
    Baseline & 4.96 & 4.8 \\ \hline
    GPT-SW3-SFT & 4.04 & 2.68 \\ \hline
    GPT-SW3-RLHF & 3.96 & 2.56 \\ \hline
    Llama2-DPO & 2.88 & 2.52 \\ \hline
    Llama2-SFT & 2.92 & 2.04 \\ \hline
    GPT-SW3-DPO & 3.24 & 1.96 \\ \hline
    \end{tabular}
\caption{\label{font-table} Average scores for five model variations on the quality of the Icelandic used and the legal accuracy after being assessed by a legal expert on summaries generated for 25 court rulings.}
\label{table:summary_scores}
\end{table}

Looking at the results in Table \ref{table:summary_scores}, both variations of GPT-SW3-1.3B that were ranked in the top two positions also achieved the highest scores for Icelandic language quality and legal accuracy. The two Llama2-7B variations exhibited similar scores for language quality, but the DPO version scored higher in legal accuracy. In contrast, the GPT-SW3-1.3B DPO variant received notably lower scores for Icelandic language quality compared to the other GPT-SW3 versions and had the lowest score for legal accuracy, despite achieving the highest ROUGE score overall. When compared to human-generated summaries, all models scored significantly lower, particularly in terms of legal accuracy.

\section{Discussion}

\subsection{Language Specific Pre-training}

After the self-supervised training on the legal text in the R dataset, the Ice-Llama2 model, which had also previously been trained on Icelandic texts from the IGC, was expected to achieve the best scores. However, the results (see Table 
\ref{table:rouge_sft_results}) indicate otherwise, showing only a marginal difference between the two Llama2-7B models. This suggests that when fine-tuning a model intended for further domain-specific training, it might be more beneficial to utilize more curated high-quality domain-specific datasets, even if this means training on less data. Such an approach allows the model to more effectively capture the relevant words and phrases it will encounter while performing downstream tasks within the specific domain, increasing the likelihood of accurately predicting the necessary tokens. Further evidence can be observed in the ROUGE score results after the summary generation training using instruction fine-tuning, where no significant difference was found between the two Llama2-7B models. The ROUGE scores for the model-generated summaries are generally modest. However, caution is needed when interpreting these results, as summaries of court rulings are often concise descriptions of the outcomes, which may not include much of the ruling's text and can be phrased differently. Consequently, assessing the quality of the generated text, based solely on N-gram overlap, can sometimes be challenging.

\subsection{Model Evaluation}

While numerical evaluations are valuable for assessing the training process, they may overlook important nuances and subjective qualities in language. This discrepancy is evident in Table \ref{tab:tbl_language_quality}, which compares the models' standings based on perplexity scores with the subjective assessments of domain experts regarding the quality of the Icelandic text generated by the models. 

\begin{table}[h]
\centering
    \begin{tabular}{|c|c|c|}
    \hline
    Rank & Perplexity Score & Qualitative Analysis \\ \hline
    1 & Llama2-DPO & GPT-SW3-SFT \\ \hline
    2 & Llama2-SFT & GPT-SW3-RLHF \\ \hline
    3 & GPT-SW3-SFT & GPT-SW3-DPO \\ \hline
    4 & GPT-SW3-RLHF & Llama2-SFT \\ \hline
    5 & GPT-SW3-DPO & Llama2-DPO \\ \hline
    \end{tabular}
\caption{Ranking of evaluated models comparing perplexity scores with results from qualitative analysis on the use of Icelandic by a domain expert.}
\label{tab:tbl_language_quality}
\end{table}

The same limitations can also be observed in in Table \ref{tab:tbl_legal_quality}, comparing the rankings of these models based on their ROUGE scores against the subjective analysis of domain experts on the legal accuracy of generated summaries:

\begin{table}[h]
\centering
    \begin{tabular}{|c|c|c|}
    \hline
    Rank & ROUGE Score & Qualitative Analysis \\ \hline
    1 & GPT-SW3-DPO & GPT-SW3-SFT \\ \hline
    2 & Llama2-DPO & GPT-SW3-RLHF \\ \hline
    3 & Llama2-SFT  & Llama2-DPO \ \\ \hline
    4 & GPT-SW3-SFT & Llama2-SFT \\ \hline
    5 & GPT-SW3-RLHF & GPT-SW3-DPO \\ \hline
    \end{tabular}
\caption{Ranking of evaluated models comparing ROUGE-scores with results from qualitative analysis on the legal accuracy in generated summaries by a domain expert.}
\label{tab:tbl_legal_quality}
\end{table}

The ROUGE scores and analyses by domain experts for both variations of the Llama2-7B model suggest that improvements can be achieved through preference training, such as DPO. In contrast, the results for the GPT-SW3-1.3B-DPO model present a different narrative. While this model demonstrates a significant improvement in ROUGE scores compared to other GPT-SW3-1.3B variants, it is frequently rated as the least preferred option by domain experts.

A detailed analysis of the legal accuracy scores reveals that the GPT-SW3-1.3B-DPO model is the only one of the model variations evaluated to receive full marks for legal accuracy in its summaries. However, it also frequently garnered low scores of 1 or 2. These contradictory results suggest that the model might have over-fitted, enabling it to sometimes produce relatively high-quality summaries while most often failing to generalize effectively.

Despite the shortcomings of DPO regarding over-fitting, its user-friendliness compared to RLHF makes it a preferable starting point for exploring whether preference training can enhance performance, as achieving stable RLHF training without divergence can present a significant challenge. Additionally, RLHF demands more computational resources than DPO. However, it cannot be overlooked, as evidenced by the results showing that the RLHF model outperformed both DPO variations in evaluations by human experts. Furthermore, RLHF offers greater flexibility in developing the reward model, as it is not limited to pairwise comparisons, which could be advantageous for specific applications.

Overall, none of the models matched the capabilities of human experts in the evaluation, especially with regard to legal accuracy. Furthermore, the discrepancy in the quality of Icelandic text between Llama2-7B and GPT-SW3-1.3B highlights the importance of language-specific pre-training, as GPT-SW3-1.3B consistently produced higher-quality text in Icelandic.

\subsection{Qualitative Analysis}

The main domain expert reviewed the output of the models  and found that they performed reasonably well overall in generating sentences that reflect the expected language and phrasing found in court rulings and summaries. However, the contextual flow between individual sentences was inconsistent, with some examples displaying a lack of cohesion and contradictory statements within the same summary, such as ``the Supreme Court dismissed the case'' and then ``the Supreme Court denied the request for dismissal of the case''. The models also struggled to adapt their summaries to the predetermined text length; with some being noticeably incomplete, while others including unnecessary sentences added to an otherwise complete summary.

While the models successfully identified essential components, such as the case subject and the court's decision, the expert found that they frequently overlooked key arguments and relevant statutes that influenced the outcome. The factual accuracy was mediocre, with several instances of contradictory statements, e.g., one correctly stating the outcome while another contradicting it. Additionally, the model occasionally confused the roles of the parties involved in a case, leading to inaccuracies about which party appealed the case or made specific claims or arguments. This sometimes carried over in the use of pronouns, creating circular sentences, such as `the claimant requested that his [own] claim be dismissed'.

This review highlights that preference training can produce legal summaries in Icelandic that are useful to some extent, but more work needs to be done before such software can be used in practice.

\section{Conclusions}

We evaluated the effect of language-specific and preference training to enhance the ability of LMs in generating Icelandic legal text summaries, compared to LMs fine-tuned solely with supervised learning.
An analysis of the evaluation results reveals that models further trained using either DPO or RLHF can exhibit improved performance in domain-specific language generation compared to those solely fine-tuned through supervised instruction; however, not consistently. Notably, this additional preference training did not lead to a general improvement in the quality of Icelandic used in the generated text. This underscores the critical role of language-specific pre-training in establishing a robust foundation for language generation.

A notable finding was the gap between ROUGE scores and expert preferences, suggesting earlier integration of human feedback could be beneficial. The dataset for pairwise comparison was based on responses with top ROUGE scores post fine-tuning. A more effective approach might involve gathering human feedback at this stage to identify which model is best suited for generating data for further training. However, this approach is constrained by the high costs associated with obtaining feedback from professional experts. While the expert feedback gathered provides valuable insight into their preferences, achieving significant improvements driven by human feedback will likely require additional resources and investment.

Future work should emphasize language-specific pre-training on Icelandic legal texts, focusing on laws, bills, and resolutions. This could enhance Icelandic quality while expanding legal knowledge. Leveraging newer models with extended context windows, such as the now available Llama3 family, could enable better utilization of training data by processing longer rulings. This capability would allow the inclusion of more training samples, potentially leading to improvements in performance. In addition to this, Retrieval-Augmented Generation (RAG) \cite{lewis2021retrievalaugmented} could be used to give models access to external knowledge when generating summaries, helping to increase factual accuracy in the responses. Moreover, a greater variety of LMs should be evaluated, as well as a larger cohort of legal experts.

\section{Limitations}
\label{sec:limitations}

A limitation of this research was the dataset size, capped at 2,600 rows, while comparable studies used about 120,000 rows \cite{stiennon2022learning}. Expanding with public court rulings and lower court summaries could improve outcomes, as a larger dataset of quality data is crucial for successfully training viable models. Additionally, the models selected were only a subset of the models available, and we had a limited number of legal experts participating in our experiments. These limitations may affect the generalization of our findings to other domains and languages.

\bibliographystyle{acl_natbib}
\bibliography{nodalida2025}

\end{document}